\documentclass[letterpaper, 10 pt, conference]{ieeeconf}  

\usepackage{blindtext}
\usepackage{graphicx}
\usepackage{subfigure}
\usepackage{caption}
\usepackage{subfig}
\usepackage{amsmath} 
\usepackage{amssymb}  
\usepackage{bm} 
\usepackage{tikz}
\usepackage{verbatim}
\usepackage{epstopdf}
\usepackage{tabularx}
\usepackage{url}
\usepackage{cite}
\usepackage{booktabs}
\usepackage{multirow}

\IEEEoverridecommandlockouts                              

\title{\LARGE \bf
	Communication constrained cloud-based long-term visual localization in real time
}

\author{Xiaqing Ding$^{1}$,
	Yue Wang$^{1}$,
	Li Tang$^{1}$,
	Huan Yin$^{1}$,
	Rong Xiong$^{1}$
	\thanks{$^{1}$Xiaqing Ding, Yue Wang, Li Tang, Huan Yin,  Rong Xiong are with the State Key Laboratory of Industrial Control and Technology, Zhejiang University, Hangzhou, P.R. China. Yue Wang is the corresponding author {\tt\small wangyue@iipc.zju.edu.cn}. Rong Xiong is the co-corresponding author {\tt\small rxiong@zju.edu.cn}.}%
}

\begin{document}

	\maketitle
	\thispagestyle{empty}
	\pagestyle{empty}

	\begin{abstract}
    Visual localization is one of the primary capabilities for mobile robots. Long-term visual localization in real time is particularly challenging, in which  the robot is required to efficiently localize itself using visual data where appearance may change significantly over time. In this paper, we  propose a cloud-based visual localization system targeting at long-term localization in real time. On the robot, we employ two estimators to achieve accurate and real-time performance.  One is a sliding-window based visual inertial odometry, which integrates constraints from consecutive observations and self-motion measurements, as well as the constraints induced by localization on the cloud. This estimator builds a local visual submap as the virtual observation which is then sent to the cloud as new localization constraints. The other one is a delayed state Extended Kalman Filter to fuse the pose of the robot localized from the cloud, the local odometry and the high-frequency inertial measurements. On the cloud, we propose a longer sliding-window based localization method  to aggregate the virtual observations for larger field of view,  leading to more robust  alignment between virtual observations and the map. Under this architecture, the robot can achieve drift-free and real-time localization using onboard resources even in a network with limited bandwidth, high latency and existence of package loss, which enables the autonomous navigation in real-world environment. We evaluate the effectiveness of our system on a dataset with challenging seasonal and illuminative variations. We further validate the robustness of the system under challenging network conditions.  
	\end{abstract}

	
	\section{Introduction}

	Localization is one of the fundamental requirements for  autonomous robots  such as driverless cars, inspection robots and service robots. 
	Visual localization is of great interest as the sensor of camera is very mature for consumer market application, which is especially favorable in places where GPS is unavailable.
	The main challenge that prevents  visual localization from deployment on the robot is the robustness against  changing appearances. Specifically, as the environment is not static in long term, the appearance can be affected by many factors, such as illumination, season and weather. Some methods have been proposed to address this challenge by improving the feature matching \cite{Jegou2010Aggregating}, promoting the search effectiveness \cite{galvez2012bags}, and introducing invariant high level semantics \cite{Naseer2017Semantics}. 
	Unfortunately, most methods are computational intensive for an on-board processing unit carried by the robot. 
	Furthermore, storing and loading of the whole map are also  unavoidable consumptions  of the resource.
	\begin{figure}[]
	\begin{center}
		\includegraphics[width=0.42\textwidth]{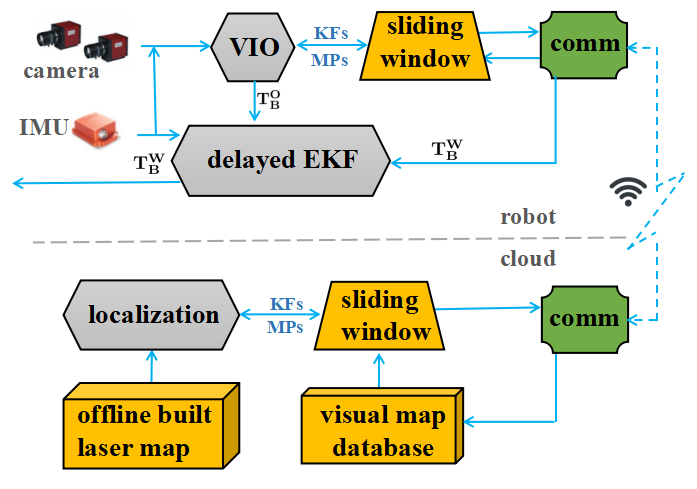}
		\caption{The framework of the  cloud-based visual inertial localization method. The method consists of a client subsystem which is applied on the robot for real-time pose estimation and a cloud subsystem designed for localization with priori laser map.  The ``KFs'' is short for ``keyframes'' and ``MPs'' is short for map points. The ``comm'' represents the communication module that is used for message transportation.}
		\label{framework}
	\end{center}
\end{figure}

	With the development of cloud computing, partial computation burden on the robot can be transferred to the cloud  to achieve robust localization in real time with limited onboard resource. Cloud robotics have demonstrated their potential in wide applications such as assembly, delivery and planning. Visual simultaneous localization and mapping (VSLAM) on the cloud  has also gained popularity recently \cite{paull2015communication,mohanarajah2015cloud}. In these works, the main focus is to build a consistent map on the cloud, while drifted position feedback on the robots  provided by visual odometry might not satisfy the requirement of autonomous navigation. There are also some  works dealing with the visual localization on the cloud \cite{middelberg2014scalable,Zhu2017Cloud}. For example in \cite{middelberg2014scalable}, the cloud receives a keyframe from the robot, estimates the global pose and sends it back to the robot.  The robot then integrates the global pose into its local visual inertial odometry (VIO) to eliminate the drift. However, when this method are applied to the long-term localization, pose estimation from only one keyframe might be unreliable due to the insufficient inliers, leading to instable  local estimation. In addition, these works rarely discuss the challenges on data transmission such as package loss, network delay and limited communication bandwidth, which could be a bottleneck of the system in the real scenario.
	
	In order to improve the robustness for localization, there is another line of methods that aggregates the local observations to generate larger field of view for localization  \cite{Milford2012SeqSLAM,Maddern2012CAT}. Although this is a promising direction for long-term applications, it may not be directly applicable to cloud robotics, as these methods use sliding-window based estimator that requires multiple frames, leading to more data transmitted through the network. 
	Therefore, the core problem for long-term localization on the cloud is the balance between the estimation quality and the network communication.
	
	In this paper, we propose a long-term localization system for cloud robotics.  To improve the robustness on localization, we consider two sliding windows in the loop to overcome the instability of the single frame estimator. To address the bottleneck on data transmission, we propose an information sparsification method to effectively transmit compressed data through the limited bandwidth communication.  We also adopt a delayed state filter to compensate for network latency and package drop.  As a result, the local estimation of robot fusing the VIO and the localization information can be drift-free in real time, which is important for autonomous navigation. In summary, the contribution of the paper can be presented as follows:
	\begin{itemize}
		\item We divide the double sliding windows estimator to the robot and the cloud to aggregate the local observations to larger virtual submap observations, which achieves more robust alignment against the map.
		\item We present an  information reduction method based on sparsification to control the data through the network, so that the requirement of the limited bandwidth can be satisfied.
		\item We build a system for long-term visual localization on the cloud that achieves the onboard drift-free pose estimation in real time, which is indispensable for autonomous navigation.
		\item We conduct experiments on real world data across seasons  to demonstrate the performance of the system in long-term localization, even under the challenging network conditions.
	\end{itemize}

	We organize the remainder of this paper as follows. In Section II some related works about long-term visual localization and cloud-based architecture will be introduced. And we give a overview of our framework in Section III. The details of the cloud-based visual localization system are demonstrated in Section IV. In Section V, the experimental results are analyzed to validate the effectiveness of the proposed method. Conclusions and some thoughts about future works are shown in Section VI.

	\section{Related works}
	
VIO uses low-cost visual sensors aided by inertial instruments to provide precise and high-frequency relative pose estimation along the robot trajectory. With the development of VIO using onboard resources, a real-time positional feedback for robot autonomous navigation becomes possible. Generally, there are two branches of methods in this area. The first branch utilizes a nonlinear filter to estimate the pose, which is very efficient and light-weighted, thus is suitable for mobile platform with limited computational resources \cite{wu2015square,mourikis2007multi}. Another branch of methods leverages non-linear optimization techniques based on local keyframes, i.e. local bundle adjustment \cite{leutenegger2015keyframe,forster2017manifold}. The optimization based methods can achieve higher performance than the filter based solution, but require more computational resources when the sliding window is long. However, the positional feedback by both branches of methods mentioned above are only reliable in short time as VIO has drift in long duration, calling for correction from the localization.

Visual localization in long term remains a open question in the community as the feature matching involving lots of outliers. Some researches resort to the more robust laser map for localization. In \cite{ding2018lasermap}, multi-session laser and visual data are used to optimize the laser map and extract the salient and stable subset for visual localization. \cite{kim2018stereo} leverages stable depth information recovered from stereo cameras for registration with prior laser map. Another way to improve localization performance in changing environment is to aggregate the variation in long term. In \cite{Paton2016Bridging,tang2019topological,churchill2013experience,burki2016appearance}, topological graph is proposed to manage experiences from multiple sessions of data, where nodes encode sensory data and edges encode relative transformations. Whenever localization fails, measurements are added to the map to enrich diversity, which helps future localization. Note that both classes of methods call for intensive computation to match the feature in a large map and solve sliding-window based optimization for more robust pose estimation.  

Instead of local features, the global image feature is shown to be more robust to appearance change. In \cite{Lowry2016Visual,Milford2012SeqSLAM}, they try to find the topological localization through place recognition and give satisfactory performance. In spite of the faster computation, this class of methods fail to provide metric localization, thus insufficient as a positional feedback for navigation.

Cloud computing provides a potential solution to real-time resource-aware SLAM and localization system, where intensive computation can be transferred to high-end servers, leaving light-weighted algorithms running on the robots. \cite{middelberg2014scalable} uses keyframes to track the pose of camera locally and sends keyframes to the server for global localization. The single frame localization suffers from the large percentage of outliers when appearance variation exists, thus brings instable estimation to the robot. \cite{mohanarajah2015cloud} also utilizes single image data to infer loop closure, thus impacted by the similar problem. In addition, the images are sent over the network in these two methods, which may require high bandwidth. Some methods designed for cooperative system demonstrate insights for the resource-aware problem. \cite{guo2018resource} proposes an efficient solution for the cooperative mapping problem by introducing augmented variables to parallelize the computation. To decrease the bandwidth requirement, \cite{paull2015communication} employs sparsification methods to compress the edges, which is also utilized in our system.

	\section{System overview}
    
    To achieve the robustness against the long-term variances, we employ a laser map to align the visual observations so that the robot can be localized \cite{ding2018lasermap}. The reason of using a laser map is that the 3D geometric structure is invariant to illuminative and perspective changes. Therefore the alignment is designed based on the geometric error between the local visual map points and the laser map points. This error is incorporated into the sliding window of the local VIO to achieve tightly coupled optimization, so that the accuracy of the laser map can be utilized to constrain the odometry. In this process, intensive computation lies on the nearest neighbor searching for data association and the nonlinear system optimization. In addition, the laser map requires memories for storage. Based on these analysis, we set to transfer the data association and the map to the cloud, and approximate the nonlinear system optimization with two sub-problems for the robot and the cloud. Note that the system presented in this paper can be applied for other type of observation models.
    
	\subsection{Notation}
	We first introduce some notations that would be used throughout the paper. We denote the  inertial measurement unit (IMU) frame as ${\mathbf{B}}$ and camera frame as ${\mathbf{C}}$. For each client robot, the coordinate of its VIO is represented as ${\mathbf{O}}$. And the coordinate of the laser map saved in the cloud is denoted as ${\mathbf{W}}$. We utilize $\mathbf{T^i_j}\in \emph{SE}(3)$ to denote the transformation between frame $\mathbf{i}$  and $\mathbf{j}$, in which the rotation matrix is $\mathbf{R^i_j}$ and the translation vector is $\mathbf{t^i_j}$. The transformation $ \mathbf{T^C_B} $ between IMU and the camera is assumed as known. The localization output at timestamp $k$ is defined as the pose of the IMU represented in the laser map $\mathbf{T^W_B}(k)$.
	
	During VIO, the state of the robot at timestamp $k$ is defined as the combination of pose $\mathbf{T^O_B}(k)$, velocity $\mathbf{v^O_B}(k)$, the acceleration measurement bias term $\mathbf{b_a}(k) \in \mathbb{R}^3$ and the  gyroscope
	measurement bias term $\mathbf{b_g}(k)\in \mathbb{R}^3$. In this paper the visual map points are denoted as $\mathbf{M_v} = \{\mathbf{p_{v\_i}}\}$ and the laser map points as $\mathbf{M_m} = \{\mathbf{p_{m\_i}}\}$.

	\subsection{Cloud-based visual inertial localization system}
    
    The whole system can be divided into the client subsystem deployed on the robot and the cloud subsystem as shown in Fig. \ref{framework}. Wireless network is used to transmit data between the two subsystem: from the client subsystem the selected states and information are sent to the cloud, and the localization result as well as the optimized states after map based optimization are sent back to the client.

    \subsubsection{Client subsystem} Client subsystem consists of a sliding-window based VIO and a delayed state Extended Kalman filter (EKF) based client-cloud state fusion estimator. The sliding-window based VIO calculates the pose $\mathbf{T^O_B}(t)$. When a new keyframe is created, its  state and the other related keyframes in the sliding-window as well as their observed visual map points are optimized. If information sent from the cloud is related to the states in the sliding-window, we also include the information within the optimization. After optimization we build a set of new edges between poses of keyframes and select the reliable subset of local visual points, to form a virtual observation, submap, which is aggregated from keyframes in the sliding window for localization on the cloud.

    The delayed state EKF based client-cloud state fusion estimator utilizes the IMU information for integration and use the localization results from the cloud as well as the local odometry results for update. The results of this filter is regarded as the output of the whole localization system on the robot. This architecture makes the localization system less sensitive to the packet loss and latency occurring in the network, and the frequency of the output is high enough as a real-time positional feedback.
	
	\begin{figure}[]
	\begin{center}
		\includegraphics[width=0.35\textwidth]{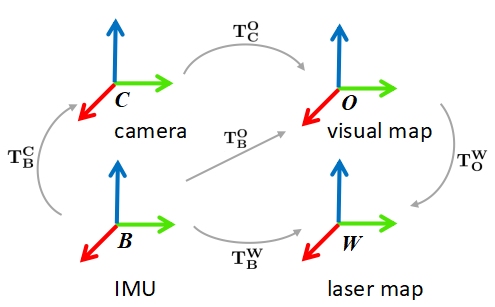}
		\caption{The coordinates and their relationship represented in this paper.}
		\label{coordination}
	\end{center}
	\end{figure}
	
	\begin{figure*}[]
	\begin{center}
		\includegraphics[width=1.0\textwidth]{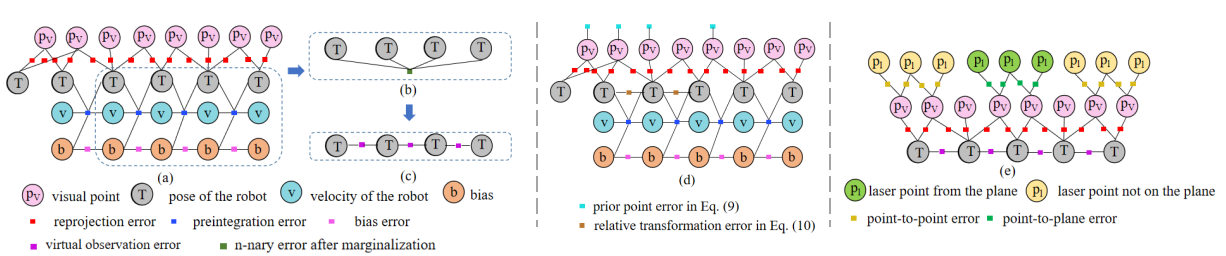}
		\caption{This figure shows the related graph models that have shown up in this paper. (a) denotes the original graph model of the inertial measurement aided bundle adjustment and the length of the sliding-window is 4 in this case. (b) demonstrates the marginalization results within the Markov Blanket (marked out with imaginary  line) after the nodes of velocity and bias are removed from the graph. And (c) denotes the graph model with the built virtual observations from the information in (b). (d) is the model used in our VIO system which merges the constraints sent from the cloud. (e) relates to the graph model in the localization optimization.}
		\label{graph_model}
	\end{center}
	\end{figure*}

	\subsubsection{Cloud subsystem} The cloud subsystem calculates the laser map aided localization optimization after receives the packets from the client, which utilizes the cloud resources to deal with data association and the numerical optimization of the large nonlinear system. The whole map is pre-loaded in the memory for further faster computation, which may not be practical for limited resources on the robot. In addition, the virtual observations sent from the robot is further aggregated in a sliding window on the cloud, so that the alignment against the map can be more robust. After optimization, the localization result as well as the optimized states that might still exist in the current sliding-window of the robot are sent back from the cloud according to the computational time and network latency.
	
	\section{System modules}

	\subsection{Sliding-window based visual inertial odometry}

	On the client robot, the sliding-window based VIO is implemented based on the architecture in \cite{mur2017visual}. The robot state is tracked in the front end. When a frame is selected as a new keyframe, the inertial measurement aided local bundle adjustment will be executed which minimizes the reprojection, preintegration and the bias error terms  related to the sliding-window. The graph model is described in Fig. \ref{graph_model}(a), in which the reprojection error for the $i$th visual point $ \mathbf{p_{v_i}} $ and $ j $th keyframe is formulated as
	\begin{equation}\label{error_reproj}
	\begin{split}
		\mathbf{e}_{reproj}(i,j)=&\rho((\pi (\mathbf{p_{v\_i},{T^O_{C_j}}})-\mathbf{u_{i,j}})^T \mathbf{\Omega}_{i,j}^{ba}\\
		&(\pi (\mathbf{p_{v\_i},{T^O_{C_j}}})-\mathbf{u_{i,j}}))
		\end{split}
	\end{equation}
	where  $ \rho(\cdot) $ is the robust kernel and $ \mathbf{\Omega} $ represents the corresponding information matrix.
	$ \mathbf{T^O_{C_j}} $ is denoted as the camera pose of the $ j $th keyframe, which can be deduced from the corresponding IMU pose $ \mathbf{T^O_{B_j}} $ and the calibration term $ \mathbf{T^C_B} $. $ \mathbf{u_{i,j}} $ represents the 2D image feature point of $\mathbf{ p_{v\_i}} $ observed by the keyframe $ j $, and $ \pi(\cdot,\cdot) $ denotes the reprojection function. The preintegration error term $ \mathbf{e}_{preint} $ and the bias error term $ \mathbf{e}_{bias} $ for keyframe $ i $ and $ k $ is defined as
	\begin{equation}\label{error_preint}
	\mathbf{e}_{preint}(i,k)=\rho([\mathbf{e}^T_R, \mathbf{e}^T_p, \mathbf{e}^T_v]\mathbf{\Omega}_{i,k}^{pre}[\mathbf{e}^T_R, \mathbf{e}^T_p, \mathbf{e}^T_v]^T)
	\end{equation}
	\begin{equation}\label{error_bias}
		\mathbf{e}_{bias}(i,k) =\rho(\mathbf{e}^T_b\mathbf{\Omega}_{i,k}^{bias}\mathbf{e}_b)
	\end{equation}
	where the error terms $ \mathbf{e}_R $,$  \mathbf{e}_p $,$ \mathbf{e}_v $ and $\mathbf{e}_b  $ are formulated as the rotational, positional, velocity and bias errors following \cite{forster2017manifold}.
	So the cost function $ \mathbf{E_{iba}} $ can be deduced
	\begin{equation}\label{cost_fuction_ba}
	\mathbf{E_{iba}} = \sum \mathbf{e}_{reproj}+ \sum \mathbf{e}_{preint} +\sum \mathbf{e}_{bias}
	\end{equation}	
	
	After optimization, the graph information is organized as a submap and sent to the cloud for localization. To decrease the demand for bandwidth, the velocity and bias nodes as well as some unstable visual points are removed from the graph.
	
	\subsubsection{Marginalize the velocity and bias nodes} The velocity and bias nodes have strong correlations with the other states in the graph. Directly marginalizing these nodes will bring in a dense information matrix that requires more bandwidth for transmission and a customized optimizer on the cloud to deal with the n-nary factor.
	Nevertheless, the Markov blanket of velocity and bias nodes does not include the visual map points. We build a set of new binary edges between the poses based on our previous work \cite{Wang2015A} which calculates the optimal representation of the marginal distribution under the divergence of Kullback-Leibler.
	
	After local bundle adjustment the optimal local  linearization  points of the states in the Markov Blanket have been estimated. We first compute the information matrix $\mathbf{\Lambda_m}$ for the graph in the Markov Blanket. For readability, here we denote $\mathbf{z_{i,j}}\in \mathbf{Z}$ as the edges between node $\mathbf{i}$ and $\mathbf{j}$ in the graph, and $\mathbf{J_{i,j}}$ as the corresponding Jacobian matrix at the linearization points. The information matrix $\mathbf{\Lambda_m}$ can be derived as
	
	\begin{equation} \label{lamba_m}
		\mathbf{\Lambda_m = \sum\limits_{z_{i,j}\in\mathbf{Z}} J^T_{i,j}\Lambda_{i,j}J_{i,j}}
	\end{equation}
	Note that we set the oldest state in the Markov Blanket as fixed so that the information matrix is invertable. By permutating the entries, we define $\mathbf{\Lambda'_m}$ as
	\begin{equation}
		\mathbf{\Lambda'_m = \left[\begin{matrix}
			\Lambda_{ss} & \Lambda_{sr}\\
			\Lambda^T_{sr} & \Lambda_{rr}
			\end{matrix}
			\right] \tag{2}}
	\end{equation}
	where the subblock related to the kept poses is $\mathbf{\Lambda_{ss}}$, and the subblock related to the removed nodes of velocity and bias is $\mathbf{\Lambda_{rr}}$. The subblock $\mathbf{\Lambda_{sr}}$ encodes the correlations between the kept poses and the removed nodes.
	The Schur complement $\mathbf{\Lambda_t}$ of the removed velocity and bias nodes is denoted as	
	\begin{equation}
		\mathbf{\Lambda_t = \Lambda_{ss}-\Lambda_{sr}(\Lambda_{rr})^{-1}\Lambda^T_{sr}}
	\end{equation}
	This information matrix is dense and can be regarded as a n-nary edge as shown in Fig. \ref{graph_model}(b). To keep the sparsity of the graph, we utilize the method proposed in \cite{Wang2015A} to build a new set of binary edges, virtual odometry, between the consecutive poses as shown in Fig. \ref{graph_model}(c). We define the Jacobian matrix of the virtual odometry measurement model linearized at the current points as $\mathbf{A}$ and the corresponding block diagonal information matrix we want to recover as $\mathbf{X}$
	\begin{equation}
		\mathbf{X = \left[
			\begin{matrix}
			X_1 & \cdots & 0\\
			\vdots & \ddots & \vdots \\
			0 & \cdots & X_l
			\end{matrix} \right]}
	\end{equation}
	in which $l$ denotes the length of the sliding window.
	Matrix $\mathbf{A}$ should be invertible as we have fixed the oldest pose in this Markov Blanket, so each subblock $\mathbf{X_i}$ in the desired information matrix $\mathbf{X}$ can be calculated  as
	\begin{equation}
		\mathbf{X_i = (\{A \Lambda^{-1}A^{T}\}_i)^{-1}}
	\end{equation}
where $\mathbf{\{\cdot\}_i}$ indicates the $i$th diagonal block with proper dimension.	

	\subsubsection{Select the robust visual map points}
	The visual map points in the sliding window might have large depth uncertainty that are not reliable to match with the laser map in 3D space. Considering the robustness of cloud-based localization and to further decrease the bandwidth requirement, we only keep those visual points that are observed more than $\theta$ times.
	
	\subsubsection{Merge information from cloud-based optimization}
	After the localization optimization from the cloud side, besides the localization results we also send the optimized relative transformations between keyframes and the positions of visual map points back to the client robot. These measurements are defined as constraints for VIO as shown in Fig. \ref{graph_model}(d). So during bundle adjustment, if a new packet of information from the cloud is received, the robot checks whether it contains information that is related to the states in current sliding-window. With this step, dynamic network latency can be considered when updating the states. If the visual map point $\mathbf{\widetilde{p}_{v\_i}}$ sent from the cloud still exists in the sliding window, its optimized position is designed as a measurement for $ \mathbf{p_{v\_i}} $ in the sliding window and we denote it as the prior point error
	\begin{equation}
	\label{error_pp}
		\mathbf{e}_{pp}(i) = \rho((\widetilde{\mathbf{p}}_{\mathbf{v\_i}}-\mathbf{p}_{\mathbf{v\_i}})^T\mathbf{\Omega}^{pt}_i(\widetilde{\mathbf{p}}_{\mathbf{v\_i}}-\mathbf{p}_{\mathbf{v\_i}})
	\end{equation}

	If keyframe $ i $ and $ j $ that relate with the relative transformation $ \mathbf{\widetilde{T}^i_j} $ sent by the cloud exist in current sliding window, $ \mathbf{\widetilde{T}^i_j} $ is utilized as a constraint for the poses of keyframe $ i $ and $ j $ and we denote it as the relative transformation error
	\begin{equation}
	\label{error_rt}
		\mathbf{e}_{rt} = \rho({Log}((\mathbf{\widetilde{T}^i_j})^{-1}\mathbf{T^i_j})^T \mathbf{\Omega}^{rt} {Log}((\mathbf{\widetilde{T}^i_j})^{-1}\mathbf{T^i_j}) )
	\end{equation}
	where $ {Log(\cdot)} $ follows the definition in \cite{Forster2015On}.

	During bundle adjustment, the error terms (\ref{error_pp}) and (\ref{error_rt}) as well as the original error terms (\ref{error_reproj}), (\ref{error_preint}) and (\ref{error_bias}) will be optimized together, based on which the precise geometry information of the laser map could be used to promote the performance of VIO. The cost function of our cloud-aided bundle adjustment now is formulated as
	\begin{equation}\label{cost_func_ibanew}
	\begin{split}
	\mathbf{E_{iba}}=&\sum \mathbf{e}_{reproj}+ \sum \mathbf{e}_{preint} +\sum \mathbf{e}_{bias}\\
	&+\sum \mathbf{e}_{pp}+\sum \mathbf{e}_{rt}
	\end{split}
	\end{equation}
	
	Note that the constraints (\ref{error_pp}) and (\ref{error_rt}) are not used to construct the information matrix $ \mathbf{\Lambda_m} $ in (\ref{lamba_m}) to avoid the information reuse.
	
	\subsection{Laser map aided localization on the cloud}
	
	The constraints of VIO are sparsified and sent from the client robot after the optimization of bundle adjustment. The information within the transmitted submap is defined with respect to the coordinate of VIO.
	The transformation $ \mathbf{T^W_O} $ between the origins of VIO and laser map is initialized at the beginning of localization as the first pose of robot with respect to the laser map. This term is used to align the states within the transmitted submap with the coordinate of laser map
	\begin{equation}
	\centering
	\begin{split}
		\mathbf{p^W_v = T^W_O p^O_v}\\
		\mathbf{T^W_B = T^W_O T^O_B}\\
		\mathbf{X = {R^W_O} X (R^W_O)^T}
	\end{split}
	\end{equation}
	
    Here we do not include the variable $ \mathbf{T^W_O} $ as a node for optimization. Instead the oldest state in the new-coming sliding-window is set as adjustable.  The point-to-point and point-to-plane constraints introduced from the laser map can be formulated as
    \begin{equation}\label{error_pt}
    \mathbf{e}_{ptp}(k,j)=\rho((\mathbf{p^W_{m\_j}-p^W_{v\_k}})^T\mathbf{\Omega}_{k,j}^{ptp}((\mathbf{p^W_{m\_j}-p^W_{v\_k}})
    \end{equation}
    \begin{equation}\label{error_pl}
    \mathbf{e}_{ptl}(k,j)=\rho((r_{n}(k,j) \mathbf{n}_{p_{m\_j}})^T \mathbf{\Omega}_{k,j}^{ptl}(r_{n}(k,j) \mathbf{n}_{p_{m\_j}}))
    \end{equation}
    where
    \begin{equation}\label{normal}
    {r}_{n}(k,j) = (\mathbf{p^W_{m\_j}-p^W_{v\_k}})^T \cdot \mathbf{n}_{p^W_{m\_j}}
    \end{equation}
    where $\mathbf{ n}_{p^W_{m\_j}} $ denotes the normal vector of laser map point $\mathbf{p^W_{m\_j}}$.
    Combined with the reprojection measurements and the virtual odometry measurements received from the robot, the cost function for localization optimization is formulated as
    \begin{equation}\label{cost_loc}
	  \mathbf{E_{loc}}=\sum \mathbf{e}_{reproj}+ \sum \mathbf{e}_{rt} +\sum \mathbf{e}_{ptp}+\sum \mathbf{e}_{ptl}
    \end{equation}
    which is represented as graph model shown in Fig. \ref{graph_model}(e).

    After optimization $ \mathbf{T^W_O} $ will be updated and saved for each client robot. Since this variable would not drift quickly, the localization could survive even it can not receive information from the client robot frequently due to the network latency or packet loss.

    The length of sliding window for VIO could not be long in order to satisfy real-time performance. When it turns to localization, the results might be trapped into local minima if the environment has changed a lot. In this cloud-based framework,  we keep a longer sliding-window to merge more environmental information in the cloud side for more accurate and robust localization.
	When a new submap is received, we will check whether the visual map points and keyframes have been registered in the map database. If the point or keyframe has been registered, the corresponding item in the database will be updated by applying new state estimation and appending new observations if possible. Or a new item will be inserted into the database. After the  update, a long sliding-window related to the latest keyframe $k$ is established for localization optimization. Then the localization result $\mathbf{T^W_{B\_k}}$ as well as the optimized points and relative transformations will be sent back to the robot. Note that to decrease the bandwidth burden, only the transformations among the latest $l$ keyframes ($l$ is the length of sliding-window in VIO) and their related visual map points will be sent to the robot.

	\subsection{Client-cloud state fusion}
	After optimization in the cloud, the localization result $ \mathbf{T^W_B} $ will be sent to the client robot at a low frequency, which could not satisfy the navigation requirement. 
	The delayed state EKF based client-cloud state fusion estimator is implemented on the robot side to merge the localization results with the local odometry for higher  localization  frequency.

	We design the state fusion module based on the multi-sensor fusion framework \cite{lynen13robust}. This module gives fusion output at the frequency of IMU. When the client robot receives the first localization result from the cloud, initialization will be called to update the information of $ \mathbf{T^W_O} $.  During the working process, the current state $ \mathbf{T^W_B}(t) $ of the robot at timestamp $t$   can be integrated from the IMU messages. And the state at timestamp $t$ could be updated  when receives odometry data  $ \mathbf{T^O_B}(t-t_o) $ from the client robot or localization data  $ \mathbf{T^W_B}(t-t_l) $ from the cloud. Note that $ \mathbf{T^W_O} $ is also formulated as a variable for estimation, which aims to compensate the drift of visual inertial odometry.
	
	\section{Experimental results}

\begin{figure}[]
	\begin{center}
		\includegraphics[width=0.42\textwidth]{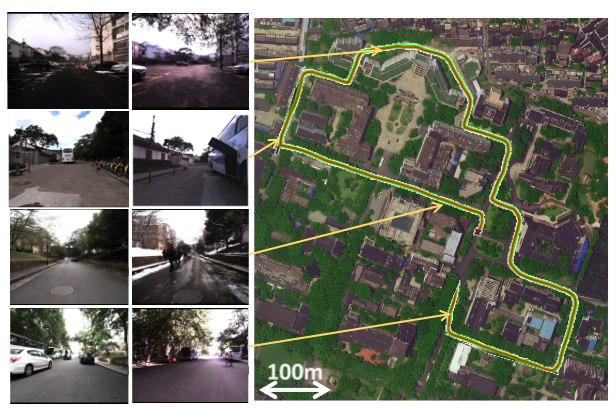}
		\caption{The satellite imagery of YQ dataset and some scenes captured at different time.}
		\label{scene}
	\end{center}
\end{figure}

	To evaluate the effectiveness of the proposed cloud-based visual inertial localization method, we test it on the real world YQ dataset that was also used for experiments in our previous work \cite{ding2018lasermap}. The dataset was collected in a campus which includes many dynamics such as the pedestrian, cyclist and significant appearance change across seasons and weather as shown in Fig. \ref{scene}. We utilize the same laser map as in \cite{ding2018lasermap} for visual localization which was built based on 21 sessions of data collected along almost the same route  within three days in early spring, 2017. The data used for localization was collected in summer and winter with a MTi 100 IMU and a pair of Pointgrey stereo camera. We also collected laser data from a  VLP-16 Velodyne LiDAR for groundtruth generation following the other visual localization methods such as \cite{sattler2018benchmarking}. We list the starting time and duration of data collection in TABLE \ref{tab:dataset}. Inside the session ``2017/08/24'' was collected along the opposite direction while the others along the same direction of the route for map building. And the session ``2018/01/29'' was collected after snow, in which the environment had changed a lot compared with the one of map building.
	
	With the development wireless communication, many low-power, wide-area (LPWA) technologies are studied to enable communications among things in wide area \cite{Sinha2017A}, which also provides powerful tools for the expanding of mobile robots. Among them the promising technologies such as narrowband internet of things (NB-IoT), have large network latency up to 10s and low bandwidth about 250KB/s. In our experiments, besides the performance under long-term changing environment, we also test the robustness under such network conditions.
	
	We utilize a low-power  laptop with an Intel i7-7500U CPU as the client, and a computer with an Intel i7-6700 CPU 3.40GHz and 16G RAM as the cloud.  All of the codes are implemented in C++.

%
\begin{table}[htbp]
	\centering
	\caption{The overview of the YQ dataset}
	\begin{tabular}{ccccccccc}
		\toprule
		\multicolumn{9}{c}{Map Building Dataset (early spring)} \\
		Day   & \multicolumn{8}{c}{Starting Time} \\
		2017/03/03 & \multicolumn{8}{c}{07:52, 09:20, 10:23, 11:48, 12:59, 14:34, 16:05, 17:38} \\
		2017/03/07 & \multicolumn{8}{c}{07:43, 09:06, 10:19, 12:40, 14:35, 16:28, 17:25, 18:07} \\
		2017/03/09 & \multicolumn{8}{c}{09:06, 10:03, 11:25, 15:06, 16:31} \\
		\midrule
		\multicolumn{9}{c}{Localization Testing Dataset (summer and winter)} \\
		\multicolumn{2}{c}{Starting Time } & \multicolumn{2}{c}{Duration} & \multicolumn{3}{c}{Starting Time } & \multicolumn{2}{c}{Duration} \\
		\multicolumn{2}{c}{2017/08/23 09:40:13} & \multicolumn{2}{c}{16:31} & \multicolumn{3}{c}{2017/08/24 09:21:41} & \multicolumn{2}{c}{13:21} \\
		\multicolumn{2}{c}{2017/08/27 15:22:11} & \multicolumn{2}{c}{17:03} & \multicolumn{3}{c}{2017/08/28 17:06:06} & \multicolumn{2}{c}{17:15} \\
		\multicolumn{2}{c}{2018/01/29 11:09:15} & \multicolumn{2}{c}{14:59} &       &       &       &       &  \\
		\bottomrule
	\end{tabular}%
	\label{tab:dataset}%
\end{table}%

\begin{table}[htbp]
	\centering
	\caption{The ATE results (m) of the localization method tested in wireless network}
	\begin{tabular}{cccccc}
		\toprule
		Sequences  & ATE& ATE & Sequences   &  ATE &ATE  \\
		& (median) & (std) & & (median) &(std)\\
		\midrule
		2017/08/23 & 0.600 & 0.334& 2017/08/24  & 0.587&0.528  \\
		2017/08/27 & 0.564 & 0.439& 2017/08/28  & 0.623 &0.354 \\
		2018/01/29 & 0.460 & 0.318 & average & 0.567 &0.395\\
		\bottomrule
	\end{tabular}%
	\label{tab:ate}%
\end{table}%

\begin{figure*}[]
	\begin{center}
		\includegraphics[width=0.95\textwidth]{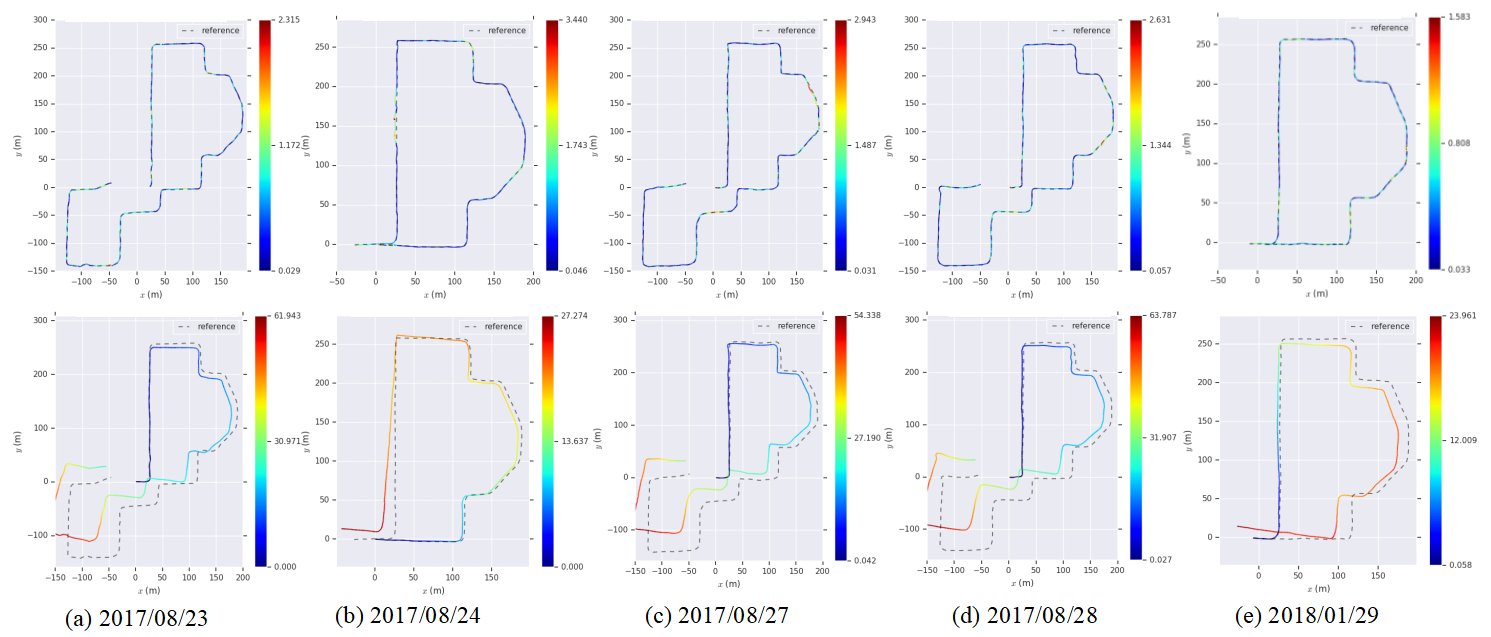}
		\caption{This figure shows the ATE errors mapped on the trajectories for each session of data. The subfigures in the first row are related to the localization results of our method, and the subfigures in the second row are related to the results of VIO.}
		\label{trajectories}
	\end{center}
\end{figure*}
	\subsection{Evaluations in the real world situation}

	We first test the effectiveness of the proposed cloud-based visual localization method in real world situation. The communication between the client and server is set up using the wireless network in the campus. The Absolute Trajectory Error (ATE) \cite{sturm2012evaluating} is utilized to evaluate the performance of the localization method. We show the results of each session of data in TABLE \ref{tab:ate} and draw the localization trajectories in the first row of Fig. \ref{trajectories}. For comparison, we also draw the trajectories of VIO in the second row of Fig. \ref{trajectories}. As the results show, the proposed cloud-based long-term visual localization method could give satisfactory results in every localization session bearing illumination, season or even heavy view point changes. Compared with VIO, our method gives the robot a long-term drift-free pose estimation, making the navigation possible. And during the whole localization process, the usage of bandwidth is around 40KB/s for transmitting the whole messages from the robot to the cloud, and around 20KB/s for transmitting from the cloud to the robot, which are much lower than the wireless connection standard.

	\subsection{Evaluations under different network conditions}

	\begin{figure}[]
		\begin{center}
			\includegraphics[width=0.45\textwidth]{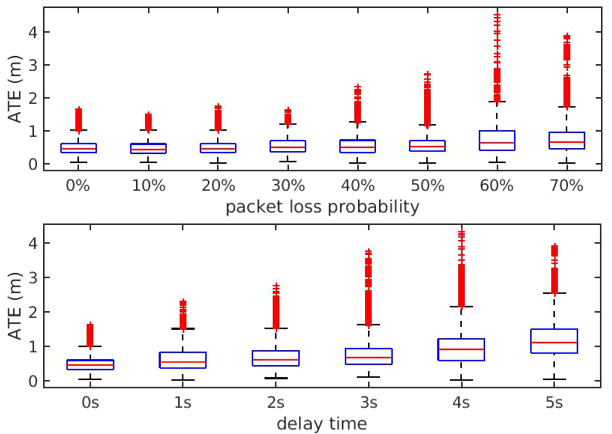}
			\caption{The localization results of session ``2018/01/29'' under different packet loss probability and network latency.}
			\label{packet loss}
		\end{center}
	\end{figure}

	Cloud-based localization method should be robust under different network conditions. In this subsection we analyze the performance of the proposed localization method when facing network latency and packet loss.
	
	To eliminate the uncertainty of the wireless network, in this subsection we utilize a cable to connect the client and server subsystem and changing the network conditions manually. We  evaluate the localization performance under different conditions of packet loss and network latency using the session of ``2018/01/29''. The localization results are shown  in Fig. \ref{packet loss} using boxplot. As we can see from the first picture in Fig. \ref{packet loss}, the localization errors do not increase much when the  packet loss probability is under 50$ \% $. And the localization method could survive even when the packet loss probability rises up to 70$\%$, which effectively validates that the proposed cloud-based visual localization method is robust to packet loss.
		\begin{figure}[]
		\begin{center}
			\includegraphics[width=0.48\textwidth]{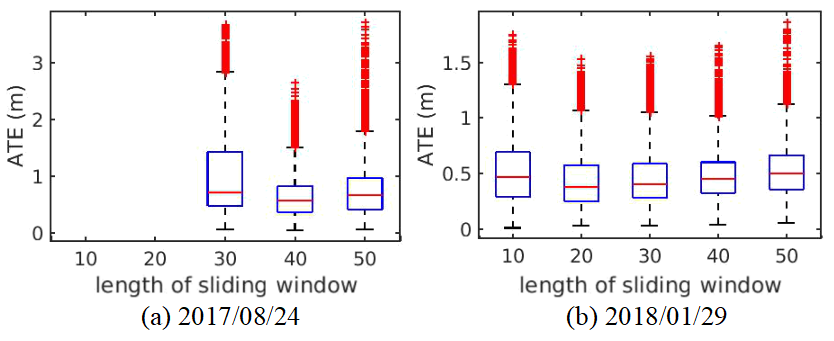}
			\caption{The localization results tested on different  lengths of the sliding window on the cloud.}
			\label{length_of_slidingwindow}
		\end{center}
	\end{figure}
	
	As for network latency, in this case we intentionally generate a network latency of the specific seconds every time the localization results are ready for transmission, which should be the worst situation the localization system would meet under the conditions of network latency. From the results we can see that the localization performance gradually deteriorates  along with the delay time. When the network latency is large, the drifted states of the EKF based state fusion localiser could not be adjusted in time, which makes the localization output drifts heavily and difficult to converge. Also large part of the optimized information sent back from the cloud can not applied on VIO as the related keyframes and map points left the sliding-window. But as the results show, our system could still give acceptable localization results when the  delay time is less than 3s, and give  rough localization results when the delay time is no more than 5s, which is more severe than lots of network condition in real world environment, thus could validate the robustness of the proposed localization method to network latency.
	
	\subsection{Effectivess of the divided double sliding-window architecture}
	We set different lengths of sliding window on the cloud side to discuss the necessity and some issues about the divided double sliding-window architecture. Since the length of sliding window is related to the observation of environment, here we utilize two sessions of data with different route directions (``2017/08/24'' and ``2018/01/29'') for experiments.
	
	The length of sliding window on the robot is 10 in our work, so the test begins at the length of 10 in the cloud side. We evaluate the localization performace with the length of 10, 20, 30, 40 and 50. The results of each session are draw in Fig. \ref{length_of_slidingwindow}. As the results show, in session ``2017/08/24'' the localiser fails to achieve localization along the whole trajectory with the length of 10 and 20. And in the two sessions the localization performances grow better at first, then gradually grow worse. In session ``2017/08/24'' the route is along the opposite direction to the one for map building. We think that the differences of the spatial distributions between the maps built along different directories influence the performance of localization, especially when the environment has changed. Aggregating more information into sliding window could relieve this problem. When the sliding window grows longer, the variable $\mathbf{T^W_O}$ might drift. However, the drift would not be taken into consideration during each optimization process in the cloud. Thus the accuracy would decrease if the sliding window is too long. So there should be a balance between accuracy and robustness. We set the length of sliding window on the cloud as 40 in other experiments in this paper.

	\section{Conclusions}
	In this paper a cloud-based visual localization method is proposed, which is designed for robots with limited resources to achieve long-term and real-time  localization. A novel robot-cloud localization architecture is proposed, which could provide high-frequency and drift-free localization estimation even in a low-bandwidth network with unpredictable conditions such as high latency and packet loss. The experimental results validate the  long-term effectiveness and also  the performances under different network conditions are discussed with quantitative analysis.
	
	In the future, we would like to study on mapping the network conditions and automatically selecting the coefficients according to the network condition.
	
	\section*{Acknowledgement}
	
	This work was supported in part by the National Key R\&D Program of China (2017YFB1300400) and in part by the National Nature Science Foundation of China (U1609210).

	\addtolength{\textheight}{-12cm}   
	



	\bibliographystyle{IEEEtran}
	\bibliography{library}

\begin{thebibliography}{10}
\providecommand{\url}[1]{#1}
\csname url@samestyle\endcsname
\providecommand{\newblock}{\relax}
\providecommand{\bibinfo}[2]{#2}
\providecommand{\BIBentrySTDinterwordspacing}{\spaceskip=0pt\relax}
\providecommand{\BIBentryALTinterwordstretchfactor}{4}
\providecommand{\BIBentryALTinterwordspacing}{\spaceskip=\fontdimen2\font plus
\BIBentryALTinterwordstretchfactor\fontdimen3\font minus
  \fontdimen4\font\relax}
\providecommand{\BIBforeignlanguage}[2]{{%
\expandafter\ifx\csname l@#1\endcsname\relax
\typeout{** WARNING: IEEEtran.bst: No hyphenation pattern has been}%
\typeout{** loaded for the language `#1'. Using the pattern for}%
\typeout{** the default language instead.}%
\else
\language=\csname l@#1\endcsname
\fi
#2}}
\providecommand{\BIBdecl}{\relax}
\BIBdecl

\bibitem{Jegou2010Aggregating}
H.~Jegou, M.~Douze, C.~Schmid, and P.~Perez, ``Aggregating local descriptors
  into a compact image representation,'' \emph{Proc Cvpr}, vol. 238, no.~6, pp.
  3304--3311, 2010.

\bibitem{galvez2012bags}
D.~G{\'a}lvez-L{\'o}pez and J.~D. Tardos, ``Bags of binary words for fast place
  recognition in image sequences,'' \emph{IEEE Transactions on Robotics},
  vol.~28, no.~5, pp. 1188--1197, 2012.

\bibitem{Naseer2017Semantics}
T.~Naseer, G.~L. Oliveira, T.~Brox, and W.~Burgard, ``Semantics-aware visual
  localization under challenging perceptual conditions,'' in \emph{IEEE
  International Conference on Robotics and Automation}, 2017.

\bibitem{paull2015communication}
L.~Paull, G.~Huang, M.~Seto, and J.~J. Leonard, ``Communication-constrained
  multi-auv cooperative slam,'' in \emph{2015 IEEE international conference on
  robotics and automation (ICRA)}.\hskip 1em plus 0.5em minus 0.4em\relax IEEE,
  2015, pp. 509--516.

\bibitem{mohanarajah2015cloud}
G.~Mohanarajah, V.~Usenko, M.~Singh, R.~D'Andrea, and M.~Waibel, ``Cloud-based
  collaborative 3d mapping in real-time with low-cost robots,'' \emph{IEEE
  Transactions on Automation Science and Engineering}, vol.~12, no.~2, pp.
  423--431, 2015.

\bibitem{middelberg2014scalable}
S.~Middelberg, T.~Sattler, O.~Untzelmann, and L.~Kobbelt, ``Scalable 6-dof
  localization on mobile devices,'' in \emph{European conference on computer
  vision}.\hskip 1em plus 0.5em minus 0.4em\relax Springer, 2014, pp. 268--283.

\bibitem{Zhu2017Cloud}
X.~Zhu, C.~Qiu, F.~Deng, S.~Pang, and Y.~Ou, ``Cloud‐based real‐time
  outsourcing localization for a ground mobile robot in large‐scale outdoor
  environments,'' \emph{Journal of Field Robotics}, vol.~34, 2017.

\bibitem{Milford2012SeqSLAM}
M.~J. Milford and G.~F. Wyeth, ``Seqslam: Visual route-based navigation for
  sunny summer days and stormy winter nights,'' in \emph{IEEE International
  Conference on Robotics and Automation}, 2012, pp. 1643--1649.

\bibitem{Maddern2012CAT}
W.~Maddern, M.~Milford, and G.~Wyeth, ``Cat-slam: Probabilistic localisation
  and mapping using a continuous appearance-based trajectory,''
  \emph{International Journal of Robotics Research}, vol.~31, no.~4, pp.
  429--451, 2012.

\bibitem{wu2015square}
K.~J. Wu, A.~M. Ahmed, G.~A. Georgiou, and S.~I. Roumeliotis, ``A square root
  inverse filter for efficient vision-aided inertial navigation on mobile
  devices,'' in \emph{2015 Robotics: Science and Systems Conference, RSS
  2015}.\hskip 1em plus 0.5em minus 0.4em\relax MIT Press Journals, 2015.

\bibitem{mourikis2007multi}
A.~I. Mourikis and S.~I. Roumeliotis, ``A multi-state constraint kalman filter
  for vision-aided inertial navigation,'' in \emph{Robotics and automation,
  2007 IEEE international conference on}.\hskip 1em plus 0.5em minus
  0.4em\relax IEEE, 2007, pp. 3565--3572.

\bibitem{leutenegger2015keyframe}
S.~Leutenegger, S.~Lynen, M.~Bosse, R.~Siegwart, and P.~Furgale,
  ``Keyframe-based visual-inertial odometry using nonlinear optimization,''
  \emph{The International Journal of Robotics Research}, vol.~34, no.~3, pp.
  314--334, 2015.

\bibitem{forster2017manifold}
C.~Forster, L.~Carlone, F.~Dellaert, and D.~Scaramuzza, ``On-manifold
  preintegration for real-time visual--inertial odometry,'' \emph{IEEE
  Transactions on Robotics}, vol.~33, no.~1, pp. 1--21, 2017.

\bibitem{ding2018lasermap}
X.~Ding, W.~Yue, D.~Li, T.~Li, and X.~Rong, ``Laser map aided visual inertial
  localization in changing environment,'' in \emph{Intelligent Robots and
  Systems (IROS), 2018 IEEE/RSJ International Conference on}.\hskip 1em plus
  0.5em minus 0.4em\relax IEEE, 2018, pp. 4794--4801.

\bibitem{kim2018stereo}
Y.~Kim, J.~Jeong, and A.~Kim, ``Stereo camera localization in 3d lidar maps,''
  in \emph{2018 IEEE/RSJ International Conference on Intelligent Robots and
  Systems (IROS)}.\hskip 1em plus 0.5em minus 0.4em\relax IEEE, 2018, pp.
  5826--5833.

\bibitem{Paton2016Bridging}
M.~Paton, K.~Mactavish, M.~Warren, and T.~D. Barfoot, ``Bridging the appearance
  gap: Multi-experience localization for long-term visual teach and repeat,''
  in \emph{Ieee/rsj International Conference on Intelligent Robots and
  Systems}, 2016, pp. 1918--1925.

\bibitem{tang2019topological}
L.~Tang, Y.~Wang, X.~Ding, H.~Yin, R.~Xiong, and S.~Huang, ``Topological
  local-metric framework for mobile robots navigation: a long term
  perspective,'' \emph{Autonomous Robots}, vol.~43, no.~1, pp. 197--211, 2019.

\bibitem{churchill2013experience}
W.~Churchill and P.~Newman, ``Experience-based navigation for long-term
  localisation,'' \emph{The International Journal of Robotics Research},
  vol.~32, no.~14, pp. 1645--1661, 2013.

\bibitem{burki2016appearance}
M.~B{\"u}rki, I.~Gilitschenski, E.~Stumm, R.~Siegwart, and J.~Nieto,
  ``Appearance-based landmark selection for efficient long-term visual
  localization,'' in \emph{2016 IEEE/RSJ International Conference on
  Intelligent Robots and Systems (IROS)}.\hskip 1em plus 0.5em minus
  0.4em\relax IEEE, 2016, pp. 4137--4143.

\bibitem{Lowry2016Visual}
S.~Lowry, N.~S¨¹nderhauf, P.~Newman, and J.~J. Leonard, ``Visual place
  recognition: A survey,'' \emph{IEEE Transactions on Robotics}, vol.~32,
  no.~1, pp. 1--19, 2016.

\bibitem{guo2018resource}
C.~X. Guo, K.~Sartipi, R.~C. DuToit, G.~A. Georgiou, R.~Li, J.~O'Leary, E.~D.
  Nerurkar, J.~A. Hesch, and S.~I. Roumeliotis, ``Resource-aware large-scale
  cooperative three-dimensional mapping using multiple mobile devices,''
  \emph{IEEE Transactions on Robotics}, no.~99, pp. 1--21, 2018.

\bibitem{mur2017visual}
R.~Mur-Artal and J.~D. Tard{\'o}s, ``Visual-inertial monocular slam with map
  reuse,'' \emph{IEEE Robotics and Automation Letters}, vol.~2, no.~2, pp.
  796--803, 2017.

\bibitem{Wang2015A}
Y.~Wang, R.~Xiong, and S.~Huang, ``A pose pruning driven solution to pose
  feature graphslam,'' \emph{Advanced Robotics}, vol.~29, no.~10, pp. 683--698,
  2015.

\bibitem{Forster2015On}
C.~Forster, L.~Carlone, F.~Dellaert, and D.~Scaramuzza, ``On-manifold
  preintegration for real-time visual-inertial odometry,'' \emph{IEEE
  Transactions on Robotics}, vol.~33, no.~1, pp. 1--21, 2015.

\bibitem{lynen13robust}
S.~Lynen, M.~Achtelik, S.~Weiss, M.~Chli, and R.~Siegwart, ``A robust and
  modular multi-sensor fusion approach applied to mav navigation,'' in
  \emph{Proc. of the IEEE/RSJ Conference on Intelligent Robots and Systems
  (IROS)}, 2013.

\bibitem{sattler2018benchmarking}
T.~Sattler, W.~Maddern, C.~Toft, A.~Torii, L.~Hammarstrand, E.~Stenborg,
  D.~Safari, M.~Okutomi, M.~Pollefeys, J.~Sivic \emph{et~al.}, ``Benchmarking
  6dof outdoor visual localization in changing conditions,'' in \emph{Proc.
  CVPR}, vol.~1, 2018.

\bibitem{Sinha2017A}
R.~S. Sinha, Y.~Wei, and S.~H. Hwang, ``A survey on lpwa technology: Lora and
  nb-iot,'' \emph{Ict Express}, vol.~3, no.~1, 2017.

\bibitem{sturm2012evaluating}
J.~Sturm, W.~Burgard, and D.~Cremers, ``Evaluating egomotion and
  structure-from-motion approaches using the tum rgb-d benchmark,'' in
  \emph{Proc. of the Workshop on Color-Depth Camera Fusion in Robotics at the
  IEEE/RJS International Conference on Intelligent Robot Systems (IROS)}, 2012.

\end{thebibliography}

\end{document}